%% file: manuscript.tex
\documentclass[letterpaper, 10 pt, conference]{IEEEtran} 

\usepackage{amsmath}
\usepackage[utf8]{inputenc}
\usepackage{graphicx}
\usepackage{subcaption}
\usepackage{multirow}
\usepackage{siunitx}
\usepackage{booktabs}
\usepackage{arydshln}
\usepackage{array}
\usepackage{tabularx}
\usepackage{adjustbox}
\usepackage{xcolor}
\usepackage{url}

\title{\LARGE \bf
Preterm infants' limb-pose estimation from depth images using convolutional neural networks}

\author{Sara Moccia$^{1,2}$, Lucia Migliorelli$^{1}$, Rocco Pietrini$^{1}$ and Emanuele Frontoni$^{1}$
\thanks{*This work was not supported by any organization}
\thanks{$^{1}$Department of Information Engineering,
        Universit\'a Politecnica delle Marche, Ancona (Italy)
        {\tt\small s.moccia@univpm.it}}%
\thanks{$^{2}$Department of Advanced Robotics, Istituto Italiano di Tencologia, Genoa (Italy)}%
}

\begin{document}

\IEEEoverridecommandlockouts
\IEEEpubid{\makebox[\columnwidth]{978-1-7281-1462-0/19/\$31.00~\copyright2019 IEEE \hfill} \hspace{\columnsep}\makebox[\columnwidth]{ }}

\maketitle


\begin{abstract}
Preterm infants' limb-pose estimation is a crucial but challenging task, which may improve patients' care and facilitate clinicians in infant's movements monitoring.
Work in the literature either provides approaches to whole-body segmentation and tracking, which, however, has poor clinical value, or retrieve a posteriori limb pose from limb segmentation, increasing computational costs and introducing inaccuracy sources. 
In this paper, we address the problem of limb-pose estimation under a different point of view. We proposed a 2D fully-convolutional neural network for roughly detecting limb joints and joint connections, followed by a regression convolutional neural network for accurate joint and joint-connection position estimation. Joints from the same limb are then connected with a maximum bipartite matching approach.
Our analysis does not require any prior modeling of infants' body structure, neither any manual interventions.
For developing and testing the proposed approach, we built a dataset of four videos (video length = 90 s) recorded with a depth sensor in a neonatal intensive care unit (NICU) during the actual clinical practice, achieving median root mean square distance [pixels] of 10.790 (right arm), 10.542 (left arm), 8.294 (right leg), 11.270 (left leg) with respect to the ground-truth limb pose.
The idea  of  estimating limb pose  directly from depth images may represent a future paradigm for addressing the problem of preterm-infants' movement monitoring and offer all possible support to clinicians in NICUs.
 \end{abstract}

\section{Introduction}
\label{sec:intro}

Preterm birth may affect infants' anatomical and functional development, leading to lifelong morbidity or, in worst-case scenario, mortality. Monitoring preterm infants is crucial to detect the onset of short- and long-term complications \cite{giuliani2016monitoring} and cribs in neonatal intensive care units (NICUs) are commonly equipped with a large variety of monitoring medical devices.

{The movement of preterm infants} is a strong clinical predictor to diagnose brain lesions~\cite{ferrari1990qualitative}, cognitive dysfunction~\cite{einspieler2016general}, sleep disorders \cite{werth2017unobtrusive} and pain \cite{heiderich2015neonatal}. Clinicians particularly pay attention to involuntary movements, consisting of asymmetrical and irregular banging of limb extremities (e.g., twitching and jerking) \cite{freymond1986energy}.
Despite being recognized as a crucial clinical task, preterm-infants' movement evaluation is merely qualitative and episodic, and mostly based on clinicians' (i) assessment at the crib side in NICUs or (ii) review of infants' video-recordings. Beside being time-consuming, this evaluation may be prone to inaccuracies due to clinicians' fatigue and susceptible to intra- and inter-clinician variability~\cite{bernhardt2011inter}.

\begin{figure}[tbp]
\centering	
\includegraphics[width=.5\textwidth]{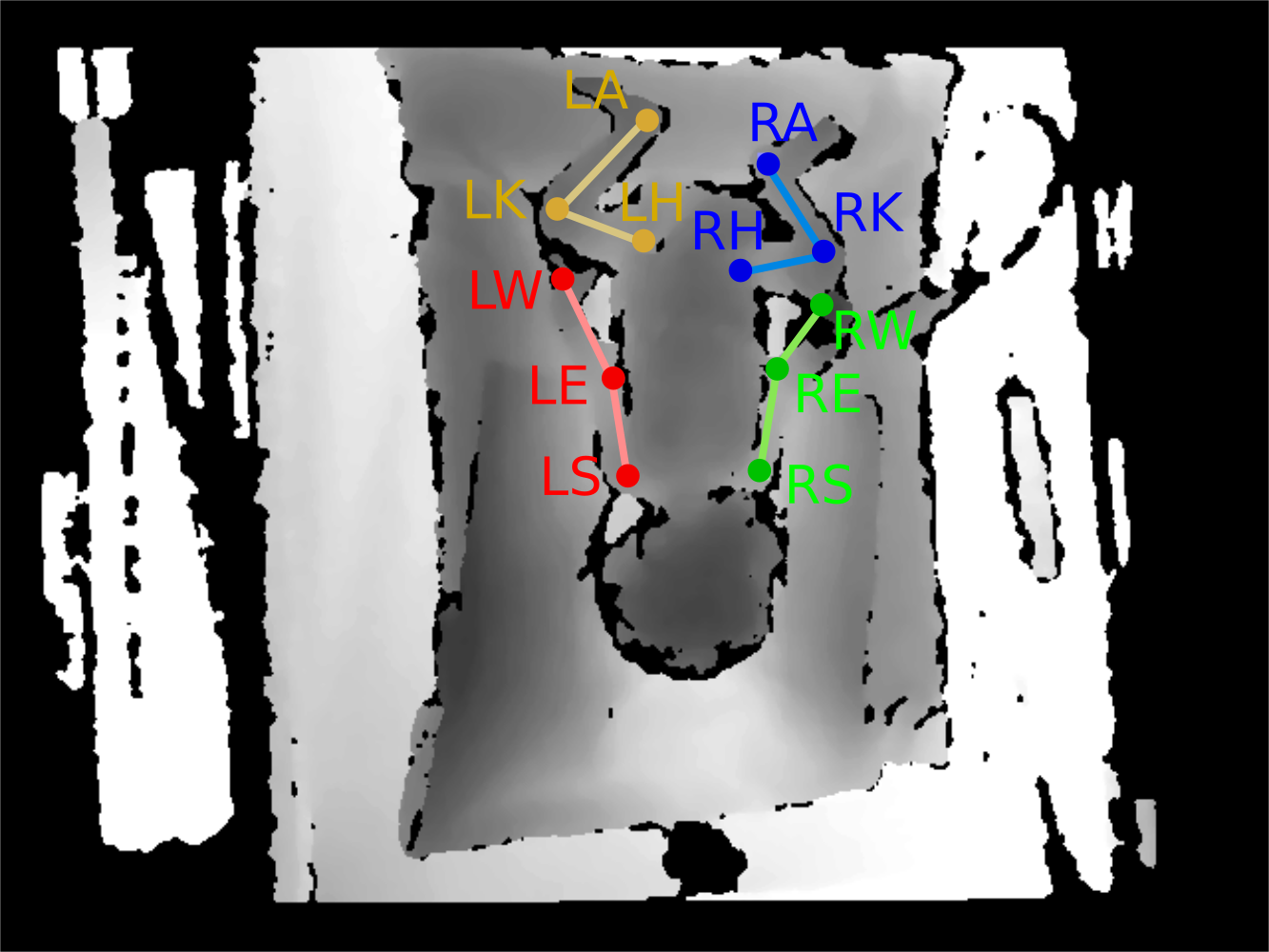}
\caption{\label{fig:joint_model} Infant model. %
LS and RS: left and right shoulder, LE and RE: left and right elbow, LW and RW: left and right wrist, LH and RH: left and right hip, LK and RK: left and right knee, LA and RA: left and right ankle.
}
\end{figure}

\begin{figure}[tbp]
\centering	
\includegraphics[width=.5\textwidth]{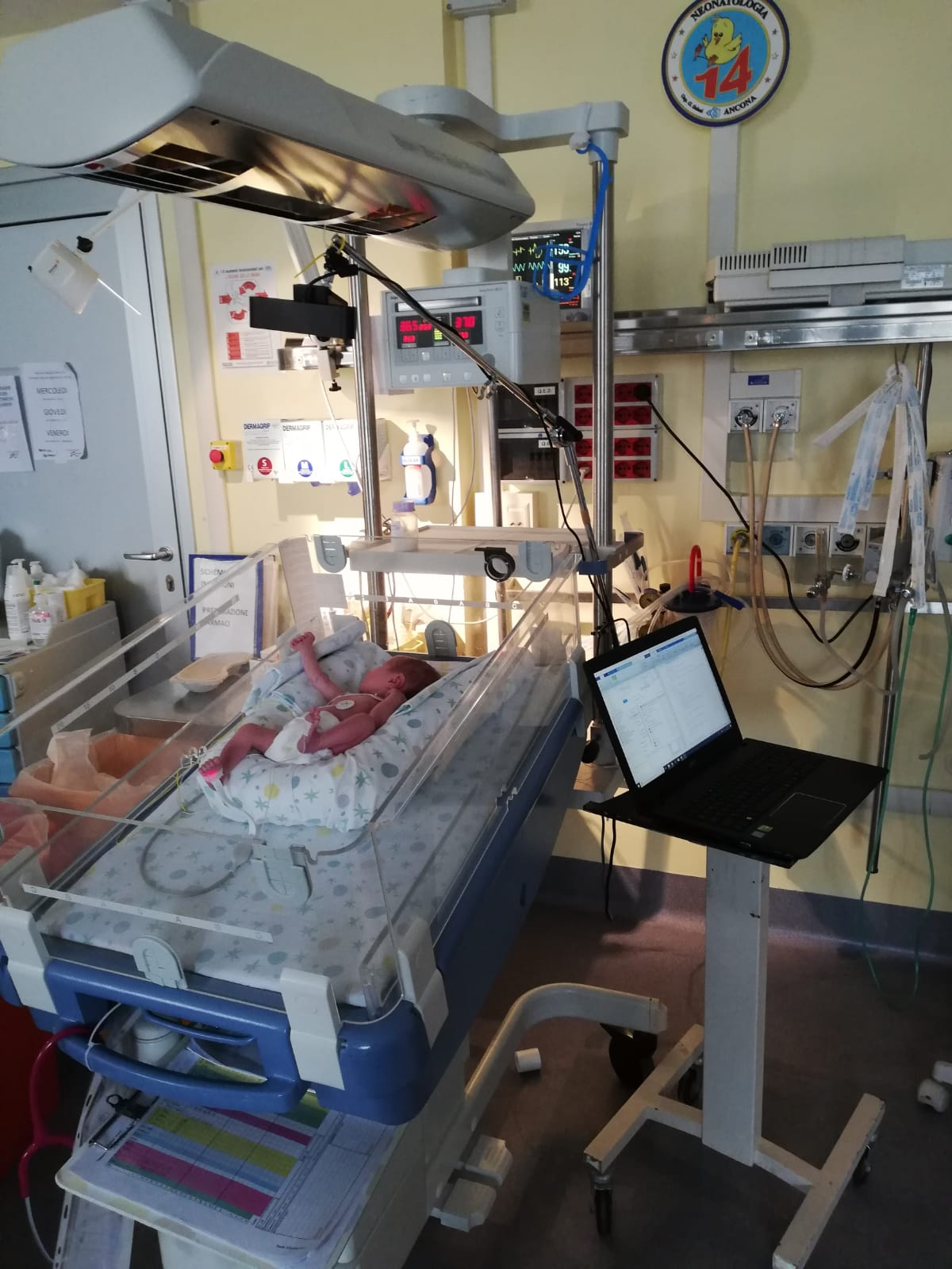}
\caption{\label{fig:system_installation} Depth-image acquisition setup. The setup does not hinder health-operator movements.
}
\end{figure}

\begin{figure*}[tbp]
\centering	
\includegraphics[width=\textwidth]{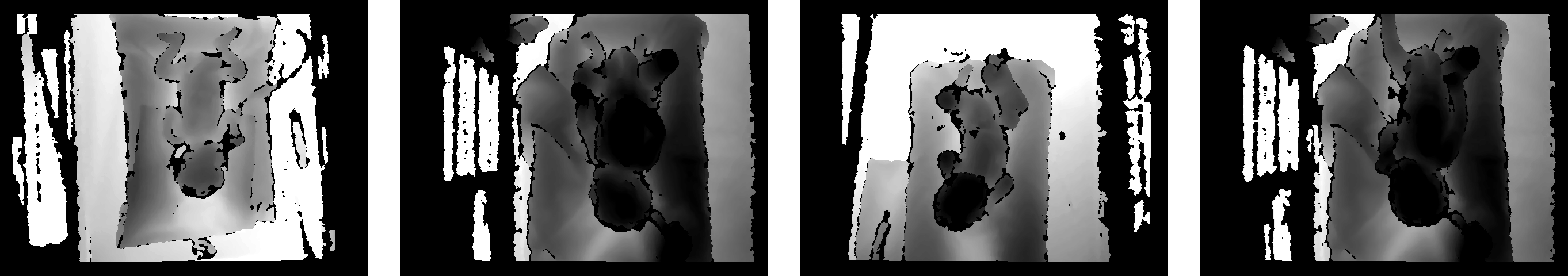}
\caption{\label{fig:dataset_challenges} Dataset challenges includes different distance between camera and infants, varying illumination level, presence of limbs self-occlusion, different number of visible joints in the camera field of view.}
\end{figure*}

Some promising computer-assisted approaches have been proposed to support clinicians in detecting infants' movement from clinical devices (e.g., accelerometer, photopletismograph and force sensors)~\cite{zuzarte2019quantifying} and multimedia data (audio and video) \cite{giuliani2016monitoring,cabon2019video,hesse2018learning}. 
With respect to intrusive clinical devices, RGB-D cameras can  be  easily  integrated  into  standard  clinical monitoring setup (e.g., over infants' cage)  while not hindering infants’ and health operators' movements.
Promising results have been achieved in the literature for whole-body detection  as a prior for infants' movement analysis.
In \cite{cenci2015non,adde2009using} threshold-based approaches to whole-body  movement detection using an RGB-D camera are proposed. In~ \cite{stahl2012optical},  optical flow and statistical classifiers are used to track manually-defined body points from RGB images.

\begin{figure*}[tbp]
\centering	
\includegraphics[width=\textwidth]{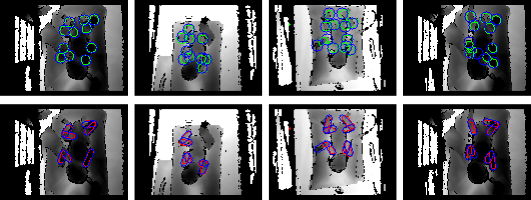}
\caption{\label{fig:detection_res}
Sample detection results. First row: ground-truth (blue) and achieved (green) joint detection. Second row: ground-truth (blue) and achieved (purple) joint-connection detection.
}
\end{figure*}

\begin{table}[tbp]
\caption{Detection-network architecture. Starting from the input depth image (1 channel), the network generates 20 maps (12 confidence maps for limb joints, and 8 affinity fields for joint connections).
}
\label{tab:archi_det}
\setlength\extrarowheight{-2.5pt}
\begin{adjustbox}{max width = .5\textwidth}
\begin{tabular}{c c c}

\textbf{Name}                   & \textbf{Kernel (Size / Stride)} & \textbf{Channels} 
\\ \hline
\multicolumn{3}{c}{\textbf{Downsampling path}}                
\\ \hline
\textbf{Input}      & --             &  1   \\ \hdashline
\textbf{Convolutional layer - Common branch}     & 3x3 / 1x1      &  64  \\ \hdashline
\textbf{Block 1 - Branch 1}     & 2x2 / 2x2     &   64  \\
                                & 3x3 / 1x1     &   64  \\
\textbf{Block 1 - Branch 2}     & 2x2 / 2x2     &   64  \\
                                & 3x3 / 1x1     &   64  \\
\textbf{Block 1 - Common branch}                & 1x1 / 1x1 & 128 \\
\hdashline
\textbf{Block 2 - Branch 1}     & 2x2 / 2x2     &   128  \\
                                & 3x3 / 1x1     &   128  \\
\textbf{Block 2 - Branch 2}     & 2x2 / 2x2     &   128  \\
                                & 3x3 / 1x1     &   128  \\
\textbf{Block 2 - Common branch}                & 1x1 / 1x1 & 256 \\

\hdashline

\textbf{Block 3 - Branch 1}     & 2x2 / 2x2     &   256  \\
                                & 3x3 / 1x1     &   256  \\
\textbf{Block 3 - Branch 2}     & 2x2 / 2x2     &   256  \\
                                & 3x3 / 1x1     &   256  \\
\textbf{Block 3 - Common branch}                & 1x1 / 1x1 & 512 \\
\hdashline
\textbf{Block 4 - Branch 1}     & 2x2 / 2x2     &   512  \\
                                & 3x3 / 1x1     &   512  \\
\textbf{Block 4 - Branch 2}     & 2x2 / 2x2     &   512  \\
                                & 3x3 / 1x1     &   512  \\
\textbf{Block 4 - Common branch}                & 1x1 / 1x1 & 1024 \\

\hline
\multicolumn{3}{c}{\textbf{Upsampling path}}                
\\ \hline
\textbf{Block 5 - Branch 1}     & 2x2 / 2x2     &   256  \\
                                & 3x3 / 1x1     &   256  \\
\textbf{Block 5 - Branch 2}     & 2x2 / 2x2     &   256  \\
                                & 3x3 / 1x1     &   256  \\
\textbf{Block 5 - Common branch}                & 1x1 / 1x1 & 512 \\
\hdashline
\textbf{Block 6 - Branch 1}     & 2x2 / 2x2     &   128  \\
                                & 3x3 / 1x1     &   128  \\
\textbf{Block 6 - Branch 2}     & 2x2 / 2x2     &   128  \\
                                & 3x3 / 1x1     &   128  \\
\textbf{Block 6 - Common branch}                & 1x1 / 1x1 & 256 \\

\hdashline
\textbf{Block 7 - Branch 1}     & 2x2 / 2x2     &   64  \\
                                & 3x3 / 1x1     &   64  \\
\textbf{Block 7 - Branch 2}     & 2x2 / 2x2     &   64  \\
                                & 3x3 / 1x1     &   64  \\
\textbf{Block 7 - Common branch}                & 1x1 / 1x1 & 128 \\

\hdashline
\textbf{Block 8 - Branch 1}     & 2x2 / 2x2     &   32  \\
                                & 3x3 / 1x1     &   32  \\
\textbf{Block 8 - Branch 2}     & 2x2 / 2x2     &   32  \\
                                & 3x3 / 1x1     &   32  \\
\textbf{Block 8 - Common branch}                & 1x1 / 1x1 & 64 \\
\hdashline
\textbf{Output}      & 1x1/1x1             &  20  \\ 
\hline
\end{tabular}
\end{adjustbox}
\end{table}

\begin{table}[tbp]
\begin{center}
\caption{Regression-network architecture. The network is fed with the depth image (1 channel) stacked with the (20) output masks of the detection network, and produces20 regression maps (12 for joints and 8 for connections).}
\label{tab:archi_reg}
\begin{tabular}{c c c}

\textbf{Name}                  & \textbf{Kernel (Size / Stride)} & \textbf{Channels} \\ \hline

\textbf{Input}             &  ---            &  21\\ 
\hdashline
\textbf{Layer 1}             & 3x3 / 1x1               & 64  \\ \hdashline
\textbf{Layer 2}           & 3x3 / 1x1                   & 128         \\ 
\hdashline
\textbf{Layer 3}              & 3x3 / 1x1                   & 256                \\ 
\hdashline
\textbf{Layer 4}              & 3x3 / 1x1       & 256                \\ 
\hdashline
\textbf{Layer 5}              & 3x3 / 1x1       & 256                \\ 
\hdashline
\textbf{Output}              & 1x1 / 1x1            &  20               \\ 
\hline
\end{tabular}
\end{center}
\end{table}

However, as explained in \cite{freymond1986energy}, single-limb movement should be evaluated to verify the presence of cerebral illnesses in preterm babies. 
An approach to limb-specific movement detection is proposed in \cite{rahmati2015weakly}. It exploits temporal tracking with particle filtering integrated with limb-trajectory priors that, however, have to be manually identified by users, hampering the usability of the approach into the actual monitoring practice. In \cite{khan2018detection}, histogram of oriented gradients is used as feature to retrieve infants' body skeleton. Body limbs and joints are a posteriori retrieved using pre-defined body-part templates.

A different strategy has been proposed in  \cite{cao2017realtime}, where  a deep-learning approach to directly assess limb joints is proposed, with advantages such as reduced computational time. 
In particular, two CNNs are used for pedestrian limb-pose estimation: the first one (a detection fully convolutional neural network, FCNN) to retrieve joint probability maps and the second one (a regression CNN) 
to refine joint-estimate position.

Inspired by \cite{cao2017realtime}, in this paper we propose to use the same strategy to estimate preterm infants' limb pose from images acquired in NICUs during the actual clinical practice.
In particular, we will focus our analysis on depth images, following recent consideration related to infants' privacy issues~\cite{hernandez2012graph,zhang2012privacy}.


%

This paper is organized as follows: Sec.~\ref{sec:meth} presents the infants' pose-estimation approach. The evaluation protocol and the image dataset built to test the proposed approach are presented in Sec.~\ref{sec:exp}. Results are presented in Sec.~\ref{sec:res} and discussed in Sec.~\ref{sec:disc}. Sec.~\ref{sec:conc} concludes this paper by summarizing the main achievements of this research.

\section{Methods}
\label{sec:meth}

Our infant's model considers each of the 4 limbs as a set of three connected joints (i.e., wrist, elbow and shoulder for arms and ankle, knee and hip for legs), as shown in Fig.~\ref{fig:joint_model}.
To estimate limb pose, we exploit two consecutive CNNs, one for detecting joints  and joint connection (Sec. \ref{sec:det}), the other for regressing the joint position, exploiting both the joint probability and joint-connection maps, with  the  latter  acting  as guidance for joint linking (Sec. \ref{sec:regr}). The joints belonging to the same limb are then connected using bipartile graph matching (Sec. \ref{sec:bipartile}).


\subsection{Detection network}
\label{sec:det}

To develop our detection FCNN, we perform multiple binary-detection operations (considering each joint and  joint-connection separately) to solve possible ambiguities of multiple joints and joint connections that may cover the same image portion (e.g. in case of limb self-occlusion). For each  video frame, we generate 20 separate ground-truth binary detection maps: 12 for the joints and 8 for the joint connections (instead of generating a single ground-truth mask with 20 different annotations, which has been shown to perform less reliably) \cite{cao2017realtime}. 
The detection network provides joint and joint-connection confidence maps as output of the joint and joint-connection branches, respectively.

For every joint mask, we consider a region of interest consisting of all pixels that lie in the circle of a given radius ($r$) centered at the joint center \cite{du2018articulated}. 
A similar approach is used to generate the ground truth for the joint connections. In this case, the ground truth is the rectangular region with thickness $r$ and centrally aligned with the joint-connection line. 

Our architecture (Table \ref{tab:archi_det}) is inspired by the classic encoder-decoder architecture of U-Net \cite{ronneberger2015u}, with 8 blocks that follows input and common-branch convolutional layers and are followed by an output layer.
Each block is divided in two branches (for joints and connections). The outputs of two branches in a block is then concatenated in a single output prior entering the next block. Using a bi-branch architecture has been shown to provide higher detection performance, as it allows processing separately the joint-probability and joint-connection affinity maps \cite{cao2017realtime}.
Batch normalization and activation with the rectified linear unit (ReLu) is performed after each convolution.

Our FCNN is trained using the per-pixel binary cross-entropy as loss function, and the adaptive moment estimation (Adam) as optimizer.

\subsection{Regression network}
\label{sec:regr}
Similarly to what is done for the detection FCNN, 
for every joint we consider a region of interest consisting of all pixels that lie in the circle with radius $r$ centered at the joint center. In this case, instead of binary masking the circle area as for the detection FCNN, we consider a Gaussian distribution with standard deviation ($\sigma$) equal to 3*$r$ and centered at the joint center. 
A similar approach is used to generate the ground-truth masks for the joint connections. In this case, the ground-truth mask is the rectangular region with thickness $r$ and centrally aligned with the joint-connection line. Pixel values in the mask are 1-D Gaussian distributed ($\sigma=3*r$) along the connection direction.

The regression network (Table \ref{tab:archi_reg}) has a single-branch architecture made of 5 layers, with an additional input and output layer. The network is fed by both the depth image and the output of the detection network, which consists of 12 joint confidence maps and 8 affinity fields for joint connections. The networks then produces 20 maps, 12 for joints and 8 for joint connections.
Batch normalization and activation with the rectified linear unit (ReLu) is performed after each convolution.

Our regression network is trained using the mean square error as loss function, and stochastic gradient descend as optimizer.

\subsection{Joint linking}
\label{sec:bipartile}
The last step of our limb pose-estimation task is to link joints for each of the infants' limb. First, we identify joint candidates from the joint regression output maps using non-maximum suppression, which is an algorithm commonly used in computer vision when redundant candidates are present \cite{hosang2017learning}. 
Once joint candidates are identified, they are linked exploiting the joint-connection regression maps. In particular, we use a bipartile matching approach, which consists in (i) computing the integral value along the line connected two candidates on the joint-connection regression output map and (ii) choosing the two winning candidates as those guaranteeing the higher integral value. 

\begin{figure}[tbp]
    \centering
    \begin{subfigure}[b]{0.5\textwidth}
        \centering
        \includegraphics[width = \textwidth]{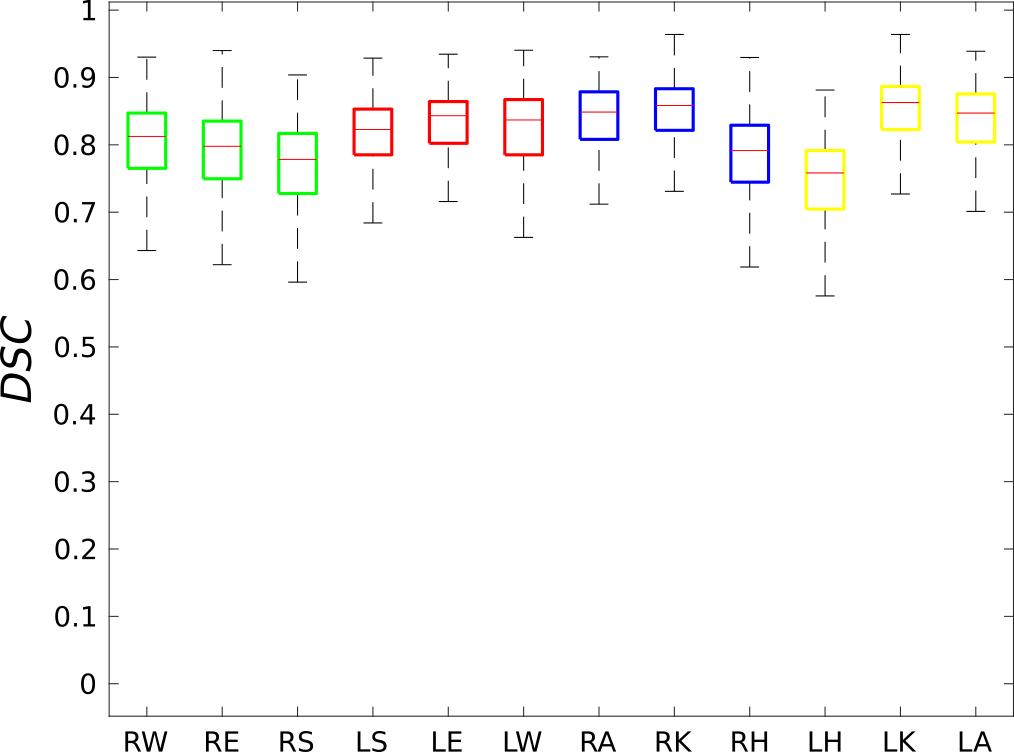}
        \caption{Joint $DSC$}
    \end{subfigure}%
    \\
    \begin{subfigure}[b]{0.5\textwidth}
        \centering
        \includegraphics[width = \textwidth]{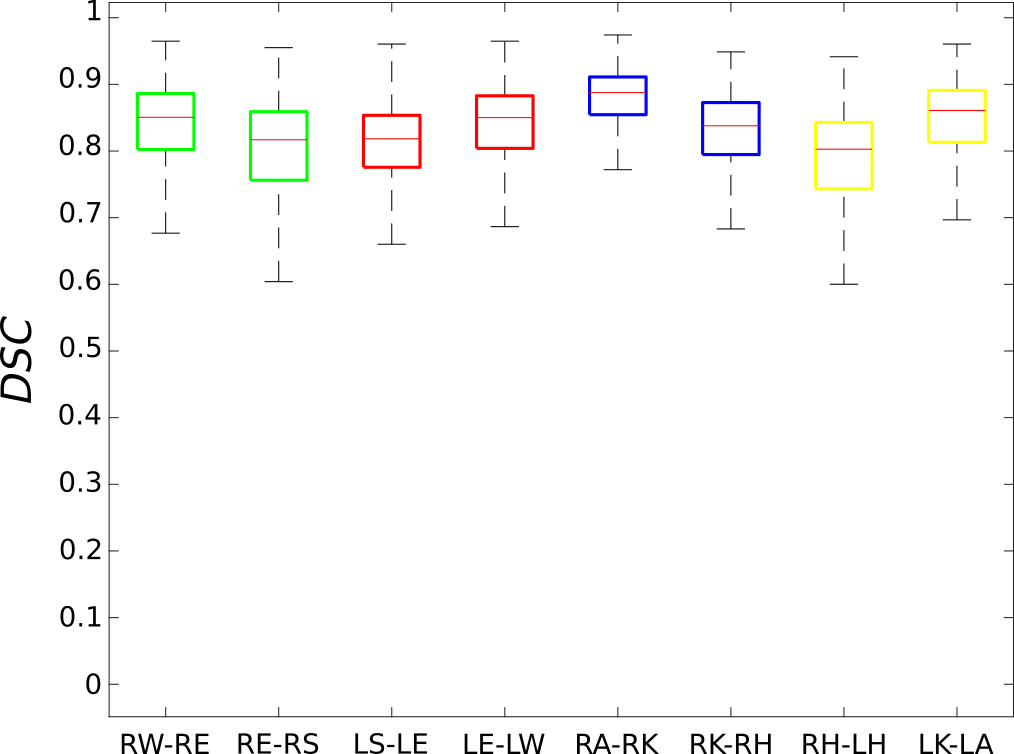}
        \caption{Connection $DSC$}
    \end{subfigure}
    \caption{\label{fig:dsc_box}Boxplots of the Dice similarity coefficient ($DSC$) for (a) joint and (b) joint-connection detection achieved with the proposed fully-convolutional neural network.}
\end{figure}

\begin{figure*}[tbp]
\centering	
\includegraphics[width=\textwidth]{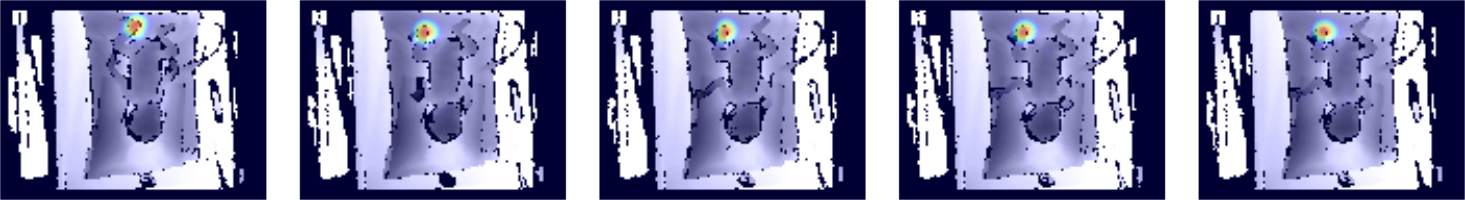}
\caption{\label{fig:regression_res}
Sample results of the regression-network output for left ankle superimposed on the corresponding depth images.
}
\end{figure*}

\section{Experimental protocol}
\label{sec:exp}

\subsection{Dataset}

Videos of four preterm infants were acquired at the G. Salesi Hospital NICU in Ancona, Italy. 
The infants were identified by clinicians in the NICU. 
All infants were spontaneously breathing and did not present hydrocephalus, congenital defects and
bronchopulmonary diseases. 
Written informed consent was
obtained from the infant's legal guardian.
Video-acquisition setup is shown in Fig. \ref{fig:system_installation}.

Video recordings (length = 90 s) were acquired for every infant, using the Astra Mini S - Orbbec~\textregistered  with a frame rate of 30 fps and image size of 640x480 pixels.
{For each video, the ground truth was manually obtained every 5 frames, resulting in 540 annotated frames per patient. 
Then, these 540 frames were split into training and testing data: 270 frames (45 s)  were used for training purpose and the remaining ones (270 frames) to test the network; resulting in a training and testing set of 1080 frames each. }

%

Challenges in the dataset included varying infant-camera distance (due to the motility of the acquisition setup), illumination level, different number of visible joints and limb self occlusion (Fig. \ref{fig:dataset_challenges}).

\subsection{Training settings}

Images were resized to 128x96 pixels in order to smooth noise and reduce both training time and  memory requirements.
Joint annotation was performed using a custom-built annotation tool, publicly available online\footnote{\url{https://github.com/roccopietrini/pyPointAnnotator}}.  
To build the ground-truth masks, we selected $r$ equal to 6 pixels.

For training the detection and regression network, we set an initial learning rate of 0.01 with a learning decay of 10\% every 10 epochs, and a momentum of 0.98. We used a batch size of 16 and for both the networks the number of epochs was set to 100.
We selected the best model as the one that maximized the accuracy on the validation set (training/validation split = 0.3).

All our analyses were performed using Keras\footnote{\url{https://keras.io/}} on a Nvidia GeForce GTX 1050 Ti/PCIe/SSE2.

\begin{figure}[tbp]
\centering	
\includegraphics[width=.5\textwidth]{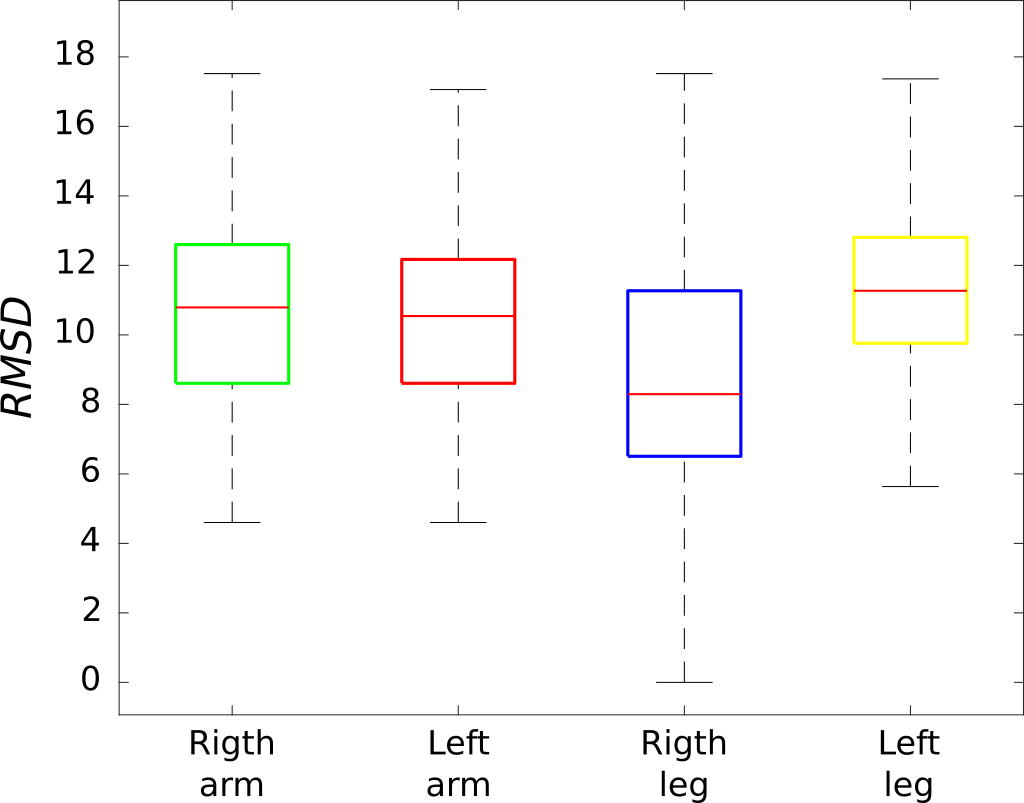}
\caption{\label{fig:pose_boxplot}Boxplots of the root mean square distance ($RMSD$) computed for the four limbs separately.}
\end{figure}

\begin{figure*}[tbp]
\centering	
\includegraphics[width=\textwidth]{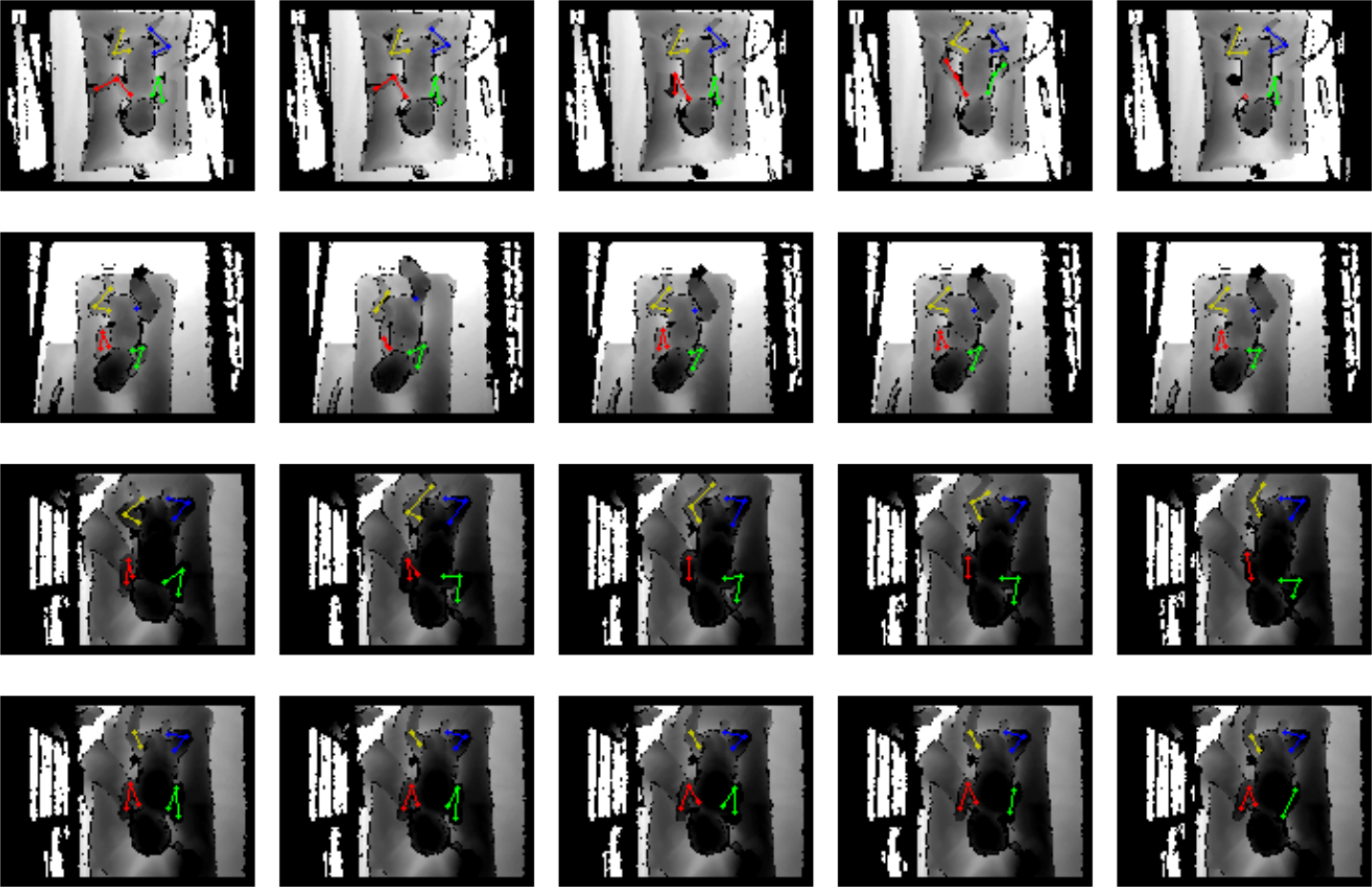}
\caption{\label{fig:pose_output}Sample pose-estimation results. Green: right-arm, red: left-arm, blue: right-leg, yellow: left-leg pose obtained with the proposed approach.}
\end{figure*}

\begin{table*}[tbp]
\centering
\caption{\label{tab:res_joint_detection}  Joint-detection performance in terms of median Dice similarity coefficient ($DSC$) and recall ($Rec$). The metrics are reported separately for each joint.}
\begin{tabular}{c|ccc|ccc|ccc|ccc}
& \multicolumn{3}{c|}{Right arm}  & \multicolumn{3}{c|}{Left arm}   &\multicolumn{3}{c|}{Right leg}  & \multicolumn{3}{c}{Left leg} \\ 
\hline
& RW & RE & RS & LS & LE & LW & RA & RK & RH & LH & LK & LA \\
\hline
$DSC$ & 0.813 & 0. 798 & 0.778 & 0.823 & 0.843 & 0.837 & 0.849 & 0.858 & 0.792 & 0.758 & 0.863 & 0.847\\
$Rec$  & 0.690 & 0. 672 & 0.637 & 0.708 & 0.734 & 0.726 & 0.752 & 0.761 & 0.664 & 0.661 & 0.770 & 0.743 \\ 
\end{tabular}
\end{table*}

\begin{table*}[tbp]
\centering
\caption{\label{tab:res_joint_connection}  Joint-connection detection performance in terms of median Dice similarity coefficient ($DSC$) and recall ($Rec$). The metrics are reported separately for each joint connection.}
\begin{tabular}{c|cc|cc|cc|cc}
& \multicolumn{2}{c|}{Right arm}  & \multicolumn{2}{c|}{Left arm}   &\multicolumn{2}{c|}{Right leg}  & \multicolumn{2}{c}{Left leg} \\ 
\hline
& RW-RE & RE-RS & LS-LE & LE-LW & RA-RK & RK-RH & RH-LH & LK-LA \\
\hline
$DSC$ & 0.851 & 0.817 & 0.818 & 0.850 &   0.888 & 0.838 &  0.803 & 0.861 \\
$Rec$ & 0.760 &  0.706 & 0.703 & 0.750 &  0.826 & 0.744 & 0.679 & 0.768\\
\end{tabular}
\end{table*}

\begin{table*}[tbp]
\centering
\caption{\label{tab:res}  Limb-pose estimation performance in terms of median root mean square distance ($RMSD$) computed with respect to ground-truth pose. The $RMSD$ is reported separately for each limb.}
\begin{tabular}{c|c|c|c|c}
& Right arm & Left arm   & Right leg  & Left leg \\ 
\hline
$RMSD$ & 10.790 & 10.542 & 8.294 & 11.270 \\
\end{tabular}
\end{table*}

\subsection{Performance metrics}
To measure the performance of the detection FCNN, we computed the Dice similarity coefficient ($DSC$) and recall ($Rec$):

\begin{equation}
DSC = \frac{2 \times TP}{2 \times TP + FP + FN}
\end{equation}
\begin{equation}
Rec = \frac{TP}{TP+FN}
\end{equation}
where $TP$: true positive, $FP$: false positive, $FN$: false negative.

To evaluate the overall pose estimation, we computed the root mean square distance ($RMSD$) [pixels] for each infants' limb.

For both the detection and regression network, we measured the testing time.

\section{Results}
\label{sec:res}

\begin{figure}[tbp]
    \centering
    \begin{subfigure}[b]{0.23\textwidth}
        \centering
        \includegraphics[width = \textwidth]{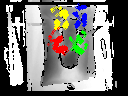}
        \caption{Joint $DSC$}
    \end{subfigure}%
    ~
    \begin{subfigure}[b]{0.23\textwidth}
        \centering
        \includegraphics[width = \textwidth]{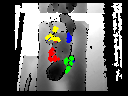}
        \caption{Connection $DSC$}
    \end{subfigure}
    \caption{\label{fig:temporal} Temporal evolution of joint position for each infants' limb. Each color refers to a different limb.}
\end{figure}

Sample outputs of the detection FCNN are shown in Fig. \ref{fig:detection_res}, both for joints and joint connections. It is worth noting that also when some joints were occluded (e.g., due to plaster as in column 2 of the image, right leg) or they were out of field of view (column 4, left leg), the network correctly detect the others.

The boxplots for $DSC$, separately computed for joints and connections, are shown in Fig. \ref{fig:dsc_box}.
Median $DSC$ and $Rec$ for joints are shown in Table \ref{tab:res_joint_detection}. For $DSC$ and $Rec$, interquartile range (IQR) was always lower than 0.080 and 0.124, respectively.
Median $DSC$ and $Rec$ for joint connections were evaluated too (Table \ref{tab:res_joint_connection}). For $DSC$ and $Rec$, IQR was always lower than 0.099 and 0.146, respectively.
Detection time was on average 0.01~s per image.

   
Visual  results for the regression-CNN output (left ankle) are shown in Fig. \ref{fig:regression_res}.
The $RMSD$ median values (for the reduced 128x96-pixel images) for pose estimation are shown in Table~\ref{tab:res}. IQR was always lower than 4.760 pixel. Boxplots for $RMSD$ are shown in Fig. \ref{fig:pose_boxplot}.

Figure~\ref{fig:pose_output} shows visual pose-estimation results for the four infants' limbs. 
Regression and bipartite-matching algorithm time was on average 0.02~s.
%
Figure \ref{fig:temporal} shows the temporal evolution of joint position for each infants' limb for two sample testing videos.

\section{Discussion}
\label{sec:disc}

The proposed FCNN achieved similar results for the detection of all joints (i.e., without outperforming in detecting one joint with respect to others), reflecting the FCNN ability of processing in parallel the different joint probability maps. This is also visible from the visual results in Fig. \ref{fig:detection_res}, where the FCNN was able to correctly detect visible joints without being affected by occluded ones.
The regression network provided guidance for the bipartile matching algorithm, which achieved satisfactory performance ($RMSD$ $<$ 12 pixels) for all limbs. 
The overall methodology required $\sim$0.03 s per image, hence being compatible with  real-time infants' monitoring.

Our approach, despite some limitations (e.g., dataset dimensions and video length),  overcame some of the literature drawbacks. Hence, it allowed to directly estimate limb-specific pose, being  computationally efficient and clinically relevant.

Future improvements of the proposed methodology may include: (i) the collection and annotation of a larger dataset (considering the lack of available datasets in this field), (ii) the analysis of temporal information (naturally encoded in videos) in both detection and regression network, as recently proposed in \cite{colleoni2019deep} and (iii) the  integration of infant-specific measures, already stored in electronic health records (e.g., height, limbs length...) to ameliorate the limb-pose estimation.
An accurate estimation may potentially allow to retrieve useful hints for movement classification (e.g., following \cite{capecci2018instrumental}) to offer all possible supports to clinicians.


\section{Conclusion}
\label{sec:conc}

In this paper, we have proposed a deep-learning framework  for  2D  pose  estimation  of  infants' limb in cages inside NICUs.  
The  framework performs first a rough detection of limb-joint position via a FCNN, and then refine the detection exploiting a regression convolutional network, followed by bipartile matching to link joints belonging to the same limb.
This work, to the best of our knowledge, represents a novel attempt to perform image-based  infants' limb pose  estimation and can  potentially  be  extended  to  handle even more complex scenario, where healthcare operators interacts with infants. 

With respect to state-of-the-art approaches, our work allows a direct estimation of limb-specific pose, is completely automatic and allows real-time processing. This make it suitable for being integrated in the clinicians' decision process and providing support for early diagnosis of brain and cognitive disorders from limb-movement analysis.

\input{manuscript.bbl}


\end{document}